\documentclass[conference]{IEEEtran}
\IEEEoverridecommandlockouts
\usepackage{cite}
\usepackage{amsmath,amssymb,amsfonts}
\usepackage{algorithmic}
\usepackage{graphicx}
\usepackage{textcomp}
\usepackage{xcolor}
\def\BibTeX{{\rm B\kern-.05em{\sc i\kern-.025em b}\kern-.08em
    T\kern-.1667em\lower.7ex\hbox{E}\kern-.125emX}}
\begin{document}

\title{Spatiotemporal deep learning model for citywide air pollution interpolation and prediction\\
}

\author{\IEEEauthorblockN{Van-Duc Le, Tien-Cuong Bui, Sang Kyun Cha}
\IEEEauthorblockA{\textit{Department of Electrical and Computer Engineering} \\
\textit{Seoul National University}\\
Seoul, Korea 08826 \\
levanduc@snu.ac.kr, cuongbt91@snu.ac.kr, chask@snu.ac.kr}
}

\maketitle

\begin{abstract}
Recently, air pollution is one of the most concerns for big cities. Predicting air quality for any regions and at any time is a critical requirement of urban citizens. However, air pollution prediction for the whole city is a challenging problem. The reason is, there are many spatiotemporal factors affecting air pollution throughout the city. Collecting as many of them could help us to forecast air pollution better. In this research, we present many spatiotemporal datasets collected over Seoul city in Korea, which is currently much suffered by air pollution problem as well. These datasets include air pollution data, meteorological data, traffic volume, average driving speed, and air pollution indexes of external areas which are known to impact Seoul's air pollution. To the best of our knowledge, traffic volume and average driving speed data are two new datasets in air pollution research. In addition, recent research in air pollution has tried to build models to interpolate and predict air pollution in the city. Nevertheless, they mostly focused on predicting air quality in discrete locations or used hand-crafted spatial and temporal features. In this paper, we propose the usage of Convolutional Long Short-Term Memory (ConvLSTM) model \cite{b16}, a combination of Convolutional Neural Networks and Long Short-Term Memory, which automatically manipulates both the spatial and temporal features of the data. Specially, we introduce how to transform the air pollution data into sequences of images which leverages the using of ConvLSTM model to interpolate and predict air quality for the entire city at the same time. We prove that our approach is suitable for spatiotemporal air pollution problems and also outperforms other related research.
\end{abstract}


\section{Introduction}
Outdoor air pollution is now threatening seriously to human health and life in big cities \cite{b8}. Many countries have constructed air pollution monitoring stations inside major cities to observe air pollutants such as PM2.5, PM10, CO, NO2, and SO2 \cite{b18}. The sources for air pollution can be from industry, people lives, vehicles, or natural sources (such as wildfires, sand storms). As a result, air pollution is affected by many complicated factors and predicting air pollution is a hard problem. 

\begin{figure}[htbp]
\centerline{\includegraphics[width=\linewidth]{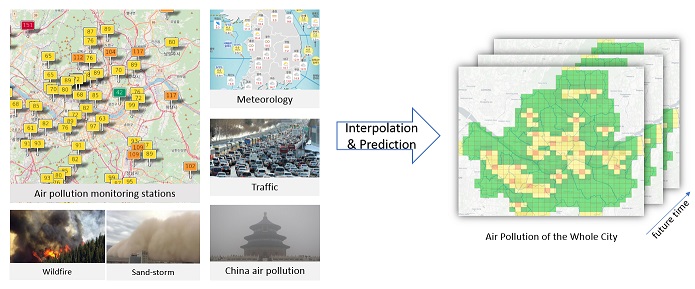}}
\caption{The overall picture of the citywide air pollution Interpolation and Prediction problem. Many factors are influencing to the air quality and we need to interpolate and predict air pollution for the whole city using all of them.}
\label{fig}
\end{figure}

Air pollution prediction has emerged as an active research field recently. Much recent research has pointed out that urban air pollution has both temporal and spatial properties as in \cite{b10,b11,b19}, to name a few. The explanation for these statements is air pollution depends on several factors both by time (temporal) and by locations (spatial). The first significant factor is meteorology, which changes in spatiotemporal form. The temperature, humidity, raining of different areas are dissimilar and the wind speed, wind direction make air pollution varies from locations to locations. Another critical reason for air pollution is traffic density and traffic congestion. The area with more traffic volume or frequent traffic jam will have ambient air quality worse. One indication of the traffic jam is the average driving speed on each road, in which a low average speed means there might be traffic congestion. The air quality monitoring stations could help us to have a measurement of air pollution at and around their located points but not for the whole city. For example, in Seoul, only less than 40 monitoring stations are covering the area of 600 km2. Consequently, we need to interpolate and predict air pollution in areas that do not have observation stations nearby. While we can not build air quality monitoring stations for all regions in a city, we can use aforementioned air pollution impact factors collected for other locations throughout the city to interpolate and forecast air pollution in the citywide scale. In fig. 1, we show the overall picture for our addressing research which tries to solve the air pollution interpolation and prediction for the entire city based on many air pollution related sources. In the next section, we will introduce the collected spatiotemporal datasets, specific to Seoul city in Korea, to help build better air pollution prediction models.

\subsection{Spatiotemporal datasets}
This section introduces the collected spatiotemporal datasets for citywide air pollution interpolation and prediction. The period of data is 3 years, from 2015 to 2017. In summary, we already recovered hourly air pollution data of 39 monitoring stations, hourly meteorological data of 28 observation stations, hourly traffic volume data for about 145 main roads, and hourly average driving speed in more than 4000 speed-surveying points in Seoul. Moreover, a recent report from \cite{b14} has shown the influence of external air pollution sources from China's cities to Seoul. To mimic these effects, we have gathered air pollution of 3 areas in China like Beijing, Shanghai, and Shandong, which affect Seoul's air quality, also from 2015 to 2017.

The hourly air pollution and meteorological datasets are quite common in recent research about air pollution predicting. The traffic volume and average driving speed data are new datasets within all known air pollution related research. The traffic volume is collected hourly by vehicle detector devices at survey points in the road. They have both the inflow and outflow directions along the survey roads. The average driving speed data is also collected hourly and by the speed checkpoints. In fig. 2, we plot the locations of all traffic volume and vehicle speed survey points in the Seoul city's map. As we can observe, the traffic volume survey points and speed checkpoints are dense and cover quite well the area of Seoul city compared with the air pollution monitoring stations (in markers). Table I also represents some statistic analysis of these 2 datasets. In Table I, turns/hr is the number of vehicles running across a survey point in an hour.

\begin{figure}[htbp]
\centerline{\includegraphics[width=\linewidth]{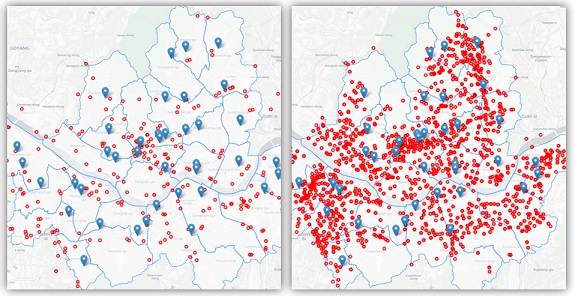}}
\caption{The location of Traffic volume survey points (in small circles) [Left] and Driving speed checkpoints (small circles) in Seoul roads [Right]. Air pollution monitoring stations in markers.}
\label{fig}
\end{figure}

\begin{table}[htbp]
\caption{Statistic of traffic volume and average driving speed data}
\begin{center}
\begin{tabular}{|c|c|c|}
    \hline & \textbf{Traffic volume} & \textbf{Average driving speed} \\ 
	\hline \textbf{count} & 4,697,888 rows & 102,453,700 rows \\ 
	\hline \textbf{mean}  & 1,510 (turns/hr) & 29.6 (km/h) \\
	\hline \textbf{min}   & 638 (turns/hr) & 0.6 (km/h)    \\ 
	\hline \textbf{max}   & 38,908 (turns/hr) & 308 (km/h)  \\ 
	\hline
\end{tabular}
\label{tab1}
\end{center}
\end{table}

Based on the observation that air pollution changes in spatiotemporal form, we propose to use a recent model named Convolutional Long Short-term Memory (ConvLSTM) by \cite{b16}. This model was proved to be superior in processing both spatial and temporal features of the input data. We also confirm in this paper that by transforming air pollution data into sequences of images, a ConvLSTM model is the best suitable for the air pollution interpolation and prediction problem, outperforms other baselines based on recurrent neural network (RNN) or convolutional neural network (CNN). Furthermore, ConvLSTM model helps us to learn the spatial and temporal features of input data at the same time and automatically, surpassing recent research that much relied on hand-crafted spatiotemporal features. We will present in some next sections how we construct the input data to fit with ConvLSTM model and how to use it efficiently in air pollution relating problems.

\section{Spatiotemporal Deep Learning Model}
In this section, we present our proposed model for citywide air pollution Interpolation and Prediction based on Spatiotemporal Deep Learning. Firstly, we briefly talk about CNN and LSTM models, which are proved is working efficiently with spatial and temporal data. Next, we propose the usage of ConvLSTM model \cite{b16} and claim its suitability for spatiotemporal air pollution problem. Finally, in the last part, we show the complete Spatiotemporal Deep Learning model for our city-scale Air Pollution Interpolation and Prediction.
\subsection{Convolutional Neural Networks (CNN) and Long Short-Term Memory (LSTM) models}
CNN is one of the most successful Deep Learning algorithms, especially in image classification, object detection. A CNN model typically consists of one or many Convolutional layers to extract the spatial relationship between input image's pixels. As a result, a CNN model can identify spatial patterns of the input such as edges, shading changes, shapes, objects, and so on. The input to a CNN is usually an image with 3 dimensions: width, height, and depth (or channel). If the image channel is 3 then we have a Red-Green-Blue (RGB) image. Alternatively, if the channel is 1, we have a gray-scale image.

LSTM is a special kind of Recurrent Neural Network (RNN), which recently works as a standard Deep Learning algorithm for sequence predicting problems like speech recognition, language translation, to name a few. The architecture of an LSTM layer is following \cite{b4} and illustrated in fig. 3.  At any time t, the input to an LSTM cell is the actual data input at time t, x\textsubscript{t}, and the hidden state from previous cell h\textsubscript{t-1}. An LSTM cell uses some "gate" mechanisms such as forget gate, input gate and output gate to decide which part of the information will be output from the cell state and which information will be stored. Following are the equations which represent the transformation from input to output of an LSTM cell. f\textsubscript{t} is the output of the forget gate, W\textsubscript{f} and b\textsubscript{f} are corresponding weights and biases. * is the matrix-vector multiplication. \(\odot\) is the Hadamard product or element-wise matrix-matrix multiplication.

\begin{equation}
    f_t = σ(W_f * [h_{t-1}, x_t] + b_f)
\end{equation}
\begin{equation}
    i_t = σ(W_i * [h_{t-1}, x_t] + b_i)
\end{equation}
\begin{equation}
    C_t = tanh(W_C * [h_{t-1}, x_t] + b_C)
\end{equation}
\begin{equation}
    C_t = f_t {\odot} C_{t-1} + i_t {\odot} \tilde{C}_t
\end{equation}
\begin{equation}
    o_t = σ(W_o * [h_{t-1}, x_t] + b_o)
\end{equation}
\begin{equation}
    h_t = o_t {\odot} tanh(C_t)
\end{equation}

The output o\textsubscript{t} in (5) and the hidden state h\textsubscript{t} in (6) is the output of the current cell, and they will be the inputs of the next cell in the LSTM loop.

\begin{figure}[htbp]
\centerline{\includegraphics[width=\linewidth]{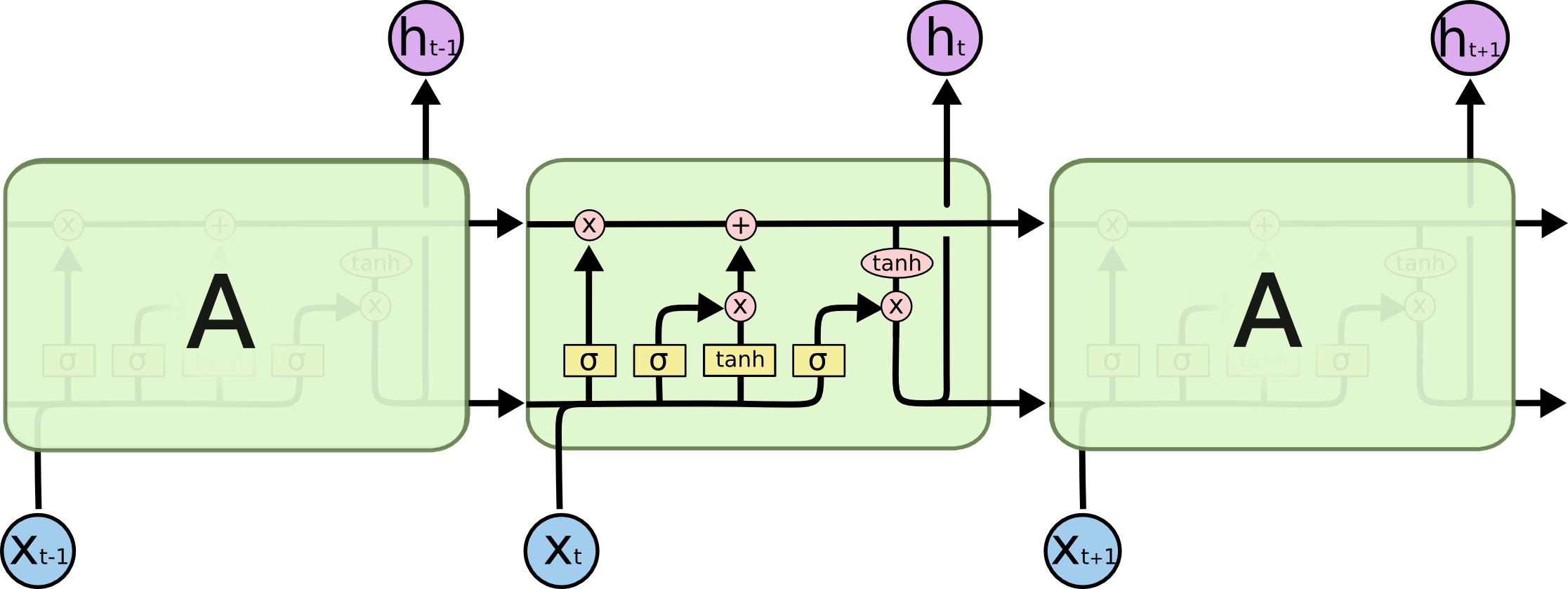}}
\caption{The architecture of a common LSTM layer with three consecutive LSTM cells. Taken from \cite{b4}.}
\label{fig}
\end{figure}

The equations from (1) to (6) are commonly applied for 1 dimensional (1D) time series input data. For a higher dimensional input such as 2D or 3D tensors data, we can easily extend these transformations by replacing matrix-vector multiplication operator with matrix-matrix multiplication. This is called fully connected LSTM (FC-LSTM) model. Nevertheless, in \cite{b16}, the authors claimed that FC-LSTM model is not efficient for spatiotemporal based data because of its poor ability in spatial learning. For the next subsection, we will describe how a new variant of LSTM model like ConvLSTM model could be fit well to our tackling problem.
\subsection{Convolutional Long Short-Term Memory (ConvLSTM)}
As presented in the sections above, urban air pollution has both spatial and temporal characteristics. Therefore, to efficiently predict air pollution anywhere (interpolation) and at any time (forecasting), we need a model that leverages both spatial and temporal features.

In 2015, X. Shi et al. proposed a model for precipitation forecasting named Convolutional LSTM Network, which was an extension of FC-LSTM model but tried to catch spatial features to have a better prediction on a spatiotemporal problem. As our air pollution problem is also spatiotemporally based, we propose to use ConvLSTM for interpolating and predicting air quality in the entire city and claim that this model gives superior performance compared with other solutions.

To address the spatiotemporal problem, in ConvLSTM model, Shi et al. proposed to replace the fully connected operators by convolutional structures in both the input-to-state and state-to-state transitions. All the inputs X\textsubscript{1},…, X\textsubscript{t}, cell outputs C\textsubscript{1},…, C\textsubscript{t}, hidden states H\textsubscript{1},…, H\textsubscript{t}, and gates i\textsubscript{t}, f\textsubscript{t}, o\textsubscript{t} of the ConvLSTM are 3D tensors whose last two dimensions are spatial dimensions. The equations for ConvLSTM are shown from (7) to (11) with * is now the convolutional operator and \(\odot\) is still element-wise matrix-matrix multiplication.

\begin{equation}
    f_t = σ(W_{xf}* X_t + W_{hf} * H_{t-1} + W_{cf} {\odot} C_{t-1} + b_f)
\end{equation}
\begin{equation}
    i_t = σ(W_{xi} * X_t + W_{hi} * H_{t-1} + W_{ci} {\odot} C_{t-1} + b_i)
\end{equation}
\begin{equation}
    C_t = f_t {\odot} C_{t-1} + i_t {\odot} tanh(W_{xc} * X_t + W_{hc} * H_{t-1} + b_c)
\end{equation}
\begin{equation}
    o_t = σ(W_{xo} * X_{t} + W_{ho} * H_{t-1} + W_{co} {\odot} C_t + b_o)
\end{equation}
\begin{equation}
    H_t = o_t {\odot} tanh(C_t)
\end{equation}

For the prediction problem, Shi et al. suggested using the structure shown in fig. 4, which consists of two networks, an encoding, and a forecasting network. The initial states and cell outputs of the forecasting network are replicated from the last state of the encoding network. Both networks are formed by stacking several ConvLSTM layers. Since the prediction target has the same dimension as the input, to generate the final prediction, all the states in the forecasting network are concatenated and fed into a 1x1 convolution layer.

\begin{figure}[htbp]
\centerline{\includegraphics[width=\linewidth]{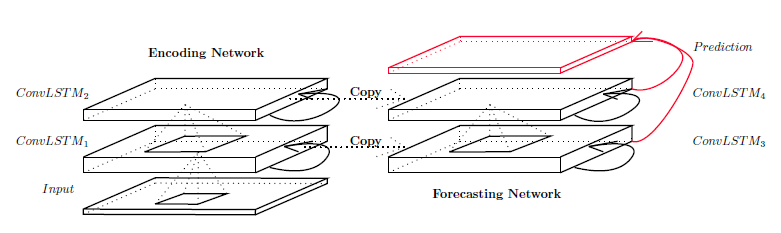}}
\caption{Encoding-Forecasting ConvLSTM structure for spatiotemporal sequence predicting, taken from \cite{b16}.}
\label{fig}
\end{figure}

\subsection{Citywide Air Pollution Interpolation and Prediction}
We need to interpolate and predict air pollution for Everywhere in a city. To leverage the ConvLSTM model, we divide the city's covering rectangle into a grid of width x height size and assign collected air pollution data into grid-cells. The value in a cell is the aggregated value of all assigned stations’ values at a timestamp t. Thus, at any time t, we have a gray-scale image of dimension width x height representing for the entire city. The pixel values are the aggregated air pollution values at that time.

Regarding the Seoul city case study, we make a 32x32 grid, which means each grid dimension has the distance approximate 1 km in the real scale. Consequently, we have many sequences of "images" which represent for the air pollution in the city by time slices. The pixel with zero value means there is no air pollution monitoring station at that grid cell. We need to predict the missing values via interpolating. As aforementioned, the air pollution in a city depends on many factors like meteorology, traffic volume, average driving speed or external air pollution sources. We also transform these data into the grid map as air pollution. For meteorological data, we assign the weather observation stations into the corresponding grid-cells and average values like in air pollution case. For traffic volume and driving speed, the survey point’s geo-locations are used to allocate them to the cell, and the traffic volume and speed are also aggregated. With external air pollution sources, because they cannot be assigned directly to the grid, we embed them into grid-map via pre-training mechanism. Fig. 5 illustrates how we construct the gray-scale images of air pollution and the spatiotemporal data for the Seoul city. We can see that, despite air pollution "images" are very sparse, other spatiotemporal data make dense images which motivate us to apply ConvLSTM model for these image sequences to interpolate and predict air pollution for the whole city.

\begin{figure}[htbp]
\centerline{\includegraphics[width=\linewidth]{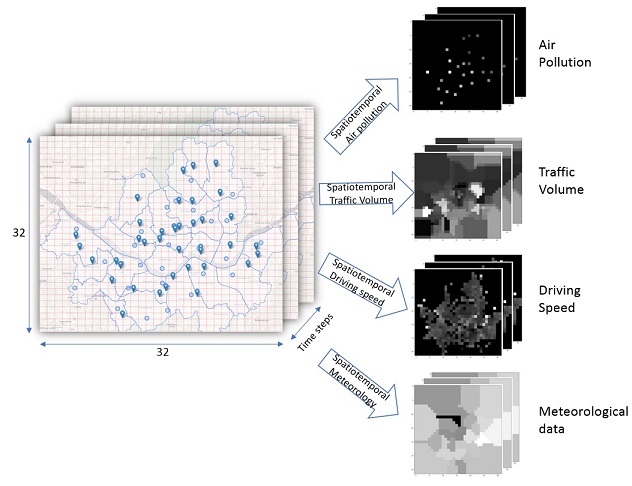}}
\caption{Gray-scale "images" of air pollution and spatiotemporal data for the Seoul city.}
\label{fig}
\end{figure}

To apply ConvLSTM model for our problem, we use gray-scale images as 2D input tensors with MxN dimension. The input tensors are not only air pollution values but the combination of air pollution and other influential factors’ values at the same location. Denotes X\textsubscript{a} \(\in\) R\textsubscript{a}\textsuperscript{P\textsubscript{a}xMxN} is the air pollution input tensor, where R\textsubscript{a} is the air pollution domain, P\textsubscript{a} is the range of air pollution values. Similarly, X\textsubscript{m} \(\in\) R\textsubscript{m}\textsuperscript{P\textsubscript{m}xMxN} is the meteorological input tensor, X\textsubscript{t} \(\in\) R\textsubscript{t}\textsuperscript{P\textsubscript{t}xMxN} is the transportation traffic input tensor, X\textsubscript{s} \(\in\) R\textsubscript{s}\textsuperscript{P\textsubscript{s}xMxN} is the vehicles average speed input tensor, and X\textsubscript{e} \(\in\) R\textsubscript{e}\textsuperscript{P\textsubscript{e}xMxN} is the external air pollution input tensor. In which R\textsubscript{m},  R\textsubscript{t}, R\textsubscript{s}, and R\textsubscript{e} are the meteorological, traffic, speed and external air pollution domain, respectively, and P\textsubscript{a}, P\textsubscript{t}, P\textsubscript{m}, and P\textsubscript{e} are the corresponding meteorological, traffic, speed and external air pollution range of values. Then the input tensor X of the model is a concatenation of all described input tensors: X = X\textsubscript{a} + X\textsubscript{m} + X\textsubscript{t} + X\textsubscript{s} + X\textsubscript{e}, in which + is a vector concatenation operator. Therefore, with our interpolation and prediction problem, if we want to forecast for K hours ahead, the equation will be following.

\begin{equation}
    \tilde{X}_{t+1},..., \tilde{X}_{t+K} = \underset{X_{t+1},..., X_{t+K}}{argmax} p(X_{t+1},..., X_{t+K} | 
                                   \hat{X}_{t-J+1},..., \hat{X}_{t})
\end{equation}

In (12), K = 1 is our interpolation and K $>$ 1 is the prediction problem. The complete model is shown in fig. 6 with 1 encoder network and 1 forecasting (decoder) network. Both 2 networks are stacks of many ConvLSTM layers. The output of the forecasting network is then fed into a 1x1 convolution layer to produce the final output. 1x1 convolution is called a “feature pooling” technique where it allows to sum pooling the features across the depth channel while still keeps the spatial characteristic of the feature map. Using 1x1 convolution at the last layer before the output layer, we can transform the ConvLSTM network’s output volume into the final output with the same 2D dimension. The output also has the grid-based form like the input, and we can use it to determine air quality's values everywhere in the city.

\begin{figure}[htbp]
\centerline{\includegraphics[width=\linewidth]{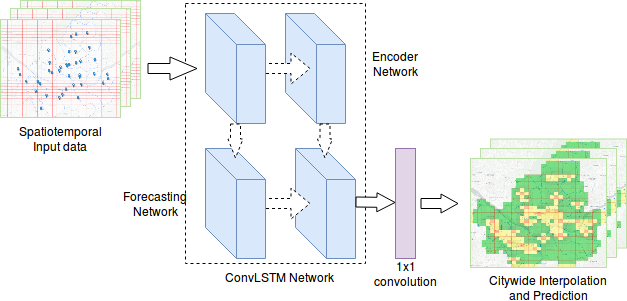}}
\caption{The Spatiotemporal Deep Learning model for Interpolating and Predicting air pollution in a Citywide scale.}
\label{fig}
\end{figure}

\section{Experiments and Evaluations}
\subsection{Baselines description}
Among recent research about interpolating and predicting air pollution at the same time, Deep Air Learning (DAL) model by \cite{b15} is the most relevant model to our approach. The authors also divided the studying city (in their case was Beijing) into the grid and tried to interpolate the air pollution in grid-cells where have no monitoring stations. Beside the interpolation capability, the authors claimed that their model was able to predict air pollution in some time ahead. The most relevant part of their research to ours is that they leveraged the using of spatial and temporal features of the input data in a unified way. Nevertheless, they still used hand-crafted spatiotemporal features for their model. On the other hand, we use ConvLSTM networks, which automatically explore the relationship of the spatial and temporal features while training with the spatiotemporal input data.

In the DAL model, they proposed a spatiotemporal semi-supervised neural network as shown in fig. 7. The authors stated that the information contained in unlabeled examples could be utilized to better exploit the geometric structure of the data, especially for the spatiotemporal data of nearby neighborhoods. Based on this characteristic, they presented a method that embeds spatiotemporal semi-supervised learning in the output layer of the neural network by minimizing the loss function between the nearby observations over the labeled and unlabeled training set, which they called spatiotemporal loss. The nearby features were chosen manually as 2 for both spatial and temporal neighbors as in their paper.

\begin{figure}[htbp]
\centerline{\includegraphics[width=\linewidth]{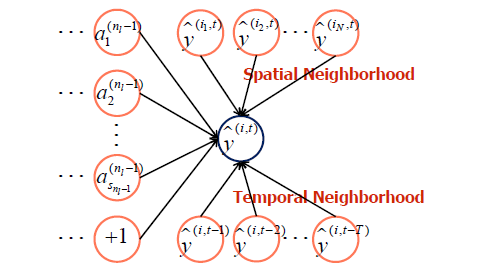}}
\caption{The graph of the spatiotemporal semi-supervised neural network in DAL model \cite{b15}.}
\label{fig}
\end{figure}

To make this model as our baseline comparison, we re-implemented it for the datasets from Seoul city. In \cite{b15}, they used a pre-trained auto-encoder for input data and then tuned with their proposed spatiotemporal loss. We also trained an auto-encoder with 4 layers and used the pre-trained model for the next phase training. We implemented DAL for both interpolation and prediction tasks.

\subsection{Data Pre-processing}
In this section, we describe how we pre-process the collected datasets for our experiments. We can refer to fig. 5 to see the results of these transformation.

The air pollution data has 6 air pollutants for each row, which are SO2, CO, O3, NO2, PM10, and PM25. Because each type of air pollutants has a different distribution, we save 6 datasets of air pollution input and train different models for each dataset with the same model architecture. For all experiments in this paper, we use PM2.5 pollutant to demonstrate our proposed model and its results. The grid-based air pollution data is then normalized to the range [0-1] by using Min-Max normalization.

Regarding meteorological data, we have 7 values like temperature, wind speed, wind direction, rainfall, lowest air pressure, highest air pressure, and humidity. To assign meteorological data into the grid cell, we aggregate the value of 5 numeric features including temperature, wind speed, air pressure, and humidity. Wind direction is a categorical feature such as North, South, West-North, and so on. Relating to wind direction, we did not average but chose one of the values if many stations are belonging to a grid cell. We tried to fill the missing meteorological values by the nearest neighbor interpolation method. The chosen interpolation method was acceptable thanks to \cite{b2}, the meteorological conditions do not change much in a range of 50 km. The resulting data is then normalized to the range [0-1] by using Min-Max normalization except for wind direction data, which is encoded by One-Hot Encoding.

The grid-based transformation for traffic volume and average driving speed is similar to air pollution. The geometric coordinates of each survey point for traffic mass and speed are used to determine its cell position in the grid-map. The data is also normalized to the range [0-1] with Min-Max normalization.

The external air pollution of 3 areas in China is kept untouched cause we use an additional neural network to embed their spatiotemporal effects with Seoul air pollution as mentioned in previous section.

For experiments and evaluations, we split the input dataset into the training set of 2 years, 2015 and 2016, and the test set is the year 2017. This splitting mechanism, choosing the training set is 2 years, and test set is the remaining year, ensures the training and test set have the same distribution and still make our model to have a good generalization. Regarding the forecasting task, we chose to predict for 12 hours ahead. That means we can predict from 1 to 12 hours in the future.

We use Tensorflow framework from Google to build ConvLSTM layers \cite{b1}. If not explicitly stated, all experiments in this paper use the learning rate is 0.001, the batch size is 128, training steps are 200, L2 regularization with beta value is 0.01 and the dropout ratio is 0.5. The metric for the test set’s result is the root mean squared error (RMSE) between the actual air pollution values and the predicted/interpolated values. This is a common metric used in the regression problem like this. RMSE is only calculated for the grid-cells that are assigned monitoring stations. If RMSE is smaller then the model’s performance is better.

\subsection{Air pollution Interpolation: Experiments and Evaluations}
We implement DAL interpolation with the output time lag is 1 hour ahead. The number of Auto-Encoder weights for each layer is 2000. After training Auto-Encoder model and save to a checkpoint, we restore the pre-trained checkpoint for spatiotemporal semi-supervised regression model training. For the spatial and temporal loss, we did not compute the loss separately for each pair of actual and prediction values but we tried to make 2 large tensors by concatenating all actual and all predicted values. Then we only need 1 computation to compute the loss for spatial or temporal neighbors. The final loss is the combination of labeled loss, weighted spatiotemporal loss of all labeled and unlabeled data. The RMSE result in the test set is shown in Table II.

The implementation of ConvLSTM interpolation model is similar with 1 hour ahead for interpolating function. The number of layers for the interpolation model is 1 encoder and 1 decoder (prediction) layer with the output channels are 64. The kernel size chosen for all layers is 3x3. We performed 2 experiments, ConvLSTM model with only labeled data loss and ConvLSTM model with both labeled and unlabeled data spatiotemporal loss as introduced in DAL model. The result is shown in Table II.

Besides DAL model as the competitive baseline, we implemented 2 other models based on Stacked FC-LSTM and CNN Encoder-Decoder to evaluate how our proposed design better in both spatial and temporal exploration, respectively. With Stacked FC-LSTM model (or FC-LSTM for short), we use the input as the gray-scale images of 32x32 size. We picked the number of hidden units for an LSTM cell is 2000 and stacked 3 LSTM cells to increase the model’s capacity. The output of LSTM cells is then flowed through a fully connected neural network (FCNN) to produce the final output. Regarding CNN Encoder-Decoder model, we applied an Encoder-Decoder network with the encoder is a convolutional layer and decoder is a deconvolution layer similar to \cite{b25}. To be comparable with ConvLSTM, we also used 1 encoder and 1 decoder layer with the same parameters (3x3 filter size and 64 output channels). The RMSE of CNN Encoder-Decoder and FC-LSTM model in the test set can be seen from Table II.

\begin{table}[htbp]
\caption{RMSE of ConvLSTM interpolation model with other baselines} 
\begin{center}
\begin{tabular}{|c|c|} 
    \hline \textbf{Interpolation models} & \textbf{RMSE} \\
	\hline \bfseries{ConvLSTM} & \bfseries{8.31466} \\
	\hline Deep Air Learing (DAL) \cite{b15} & 11.77393 \\ 
	\hline CNN Encoder-Decoder & 9.42967  \\ 
	\hline Stacked FC-LSTM & 12.01648  \\ 
	\hline \bfseries{ConvLSTM + Spatiotemporal Loss} & \bfseries{8.09817} \\
	\hline
\end{tabular}
\label{tab1}
\end{center}
\end{table}

From Table II, it is clear that ConvLSTM interpolation model achieves the best RMSE among other baselines. Moreover, ConvLSTM model with spatiotemporal loss has better RMSE than pure ConvLSTM. It infers that spatiotemporal loss is a good improvement for our addressing air pollution problem.

\subsubsection{More Interpolation Evaluation}
The most critical evaluation for this part is to evaluate the citywide air pollution Interpolation. It means how well the predicted output image reflects the air pollution of the whole city. In this part, we describe how to asset this result efficiently.

The RMSE values shown in Table II are useful for quantitative evaluation but they are hard to show the quality of interpolated results for the whole city. In fig. 8, we plot the output images of DAL, ConvLSTM, CNN Encoder-Decoder, and FC-LSTM model to see the distribution of air pollution interpolated values. Intuitively, FC-LSTM model shows the worst output with all the grid-cells except the existing monitoring stations have the same value. The reason is FC-LSTM networks do not learn the spatial features well, and thus do not give good interpolation output. For the remaining 3 models, ConvLSTM and DAL show pretty good air pollution interpolation compared with the CNN Encoder-Decoder model. To prove that ConvLSTM model produces better air pollution interpolated values over other baselines, we compare it with the actual air pollution values distribution. Here, we suggest testing the variance of the interpolated values.

Variance is the expectation of the squared deviation of a distribution from its mean. A high variance indicates that the data points are very spread out from the mean and other points. While a small variance states that the data points tend to be close to each other. We calculate the variance of actual air pollution values and the interpolated results, repeat for 10 interpolated time steps and draw to a graph in fig. 9. We can witness that the variance of interpolation values of ConvLSTM model is the closest to the variance of actual air pollution values. It means ConvLSTM interpolation model outcomes DAL and other baseline models in producing better interpolated values.

\begin{figure}[htbp]
\centerline{\includegraphics[width=\linewidth]{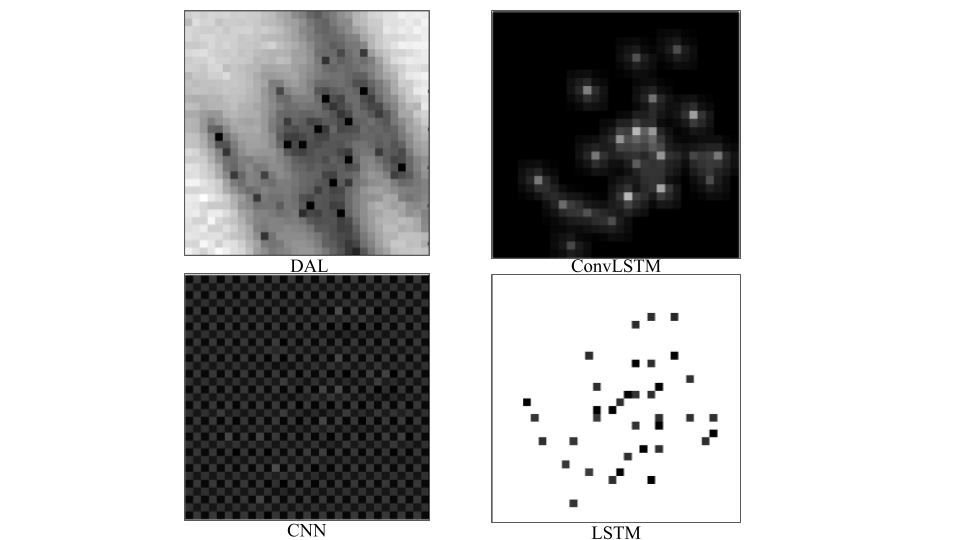}}
\caption{Plotting of interpolated output images for 4 interpolation models.}
\label{fig}
\end{figure}

\begin{figure}[htbp]
\centerline{\includegraphics[width=\linewidth]{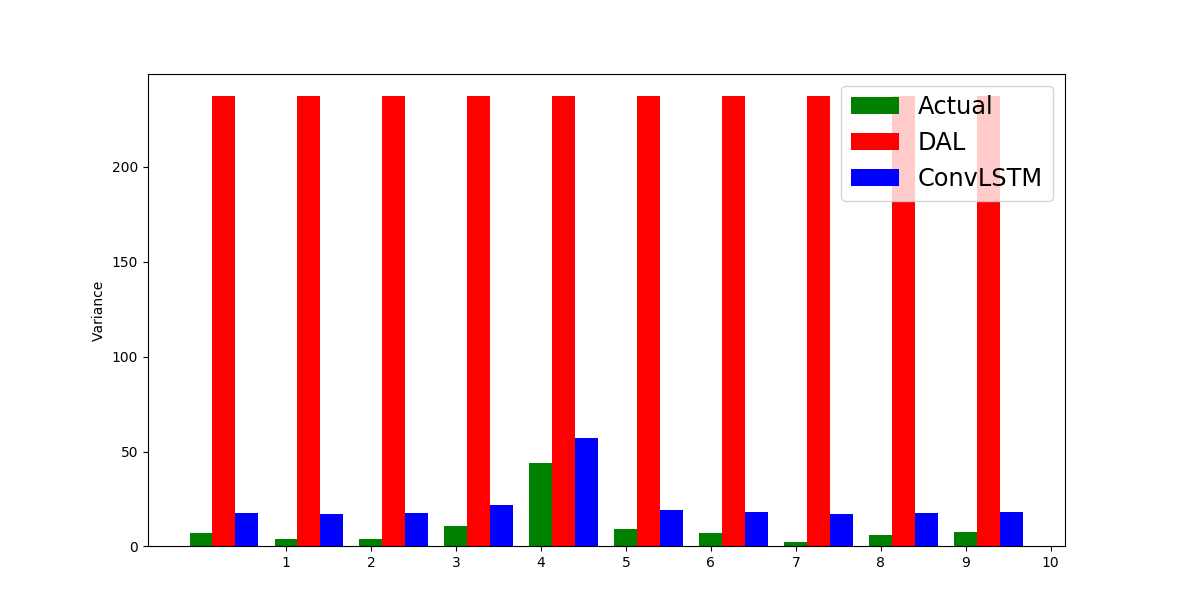}}
\caption{The Variance of actual air pollution values and interpolated values of ConvLSTM and DAL model.}
\label{fig}
\end{figure}

\subsubsection{Interpolation with air pollution influence factors}
In this section, we show the experiments’ results of air pollution interpolating along with spatiotemporal air pollution impact factors like meteorology, traffic volume, driving average speed, and external air pollution sources. We implement the experiments with following models: ConvLSTM (ConvLSTM with only air pollution data), ConvLSTM + Met (air pollution and meteorological data), ConvLSTM + Traffic (air pollution and traffic volume data), ConvLSTM + Speed (air pollution and vehicles average speed data), ConvLSTM + External (air pollution and external air pollution data), ConvLSTM + All (air pollution and all related factors). Table III shows the RMSE in the test set of pure ConvLSTM and other combination models.

\begin{table}[htbp]
\caption{Interpolation error (RMSE) of ConvLSTM model and its combination with other spatiotemporal factors} 
\begin{center}
\begin{tabular}{|c|c|c|} 
    \hline \textbf{Models} & \textbf{RMSE} & \textbf{spRMSE} \\
	\hline ConvLSTM (baseline) & 8.31466 & 15.48715 \\ 
	\hline \bfseries{ConvLSTM + Met} & \bfseries{6.58092} & 14.40496 \\
	\hline ConvLSTM + Traffic & 8.30858 & 15.47893 \\ 
	\hline ConvLSTM + Speed & 8.91373 & 15.17757 \\ 
	\hline ConvLSTM + External & 6.63926 & 14.46107 \\ 
	\hline \bfseries{ConvLSTM + All} & 7.17028 & \bfseries{11.02544}  \\ 
	\hline
\end{tabular}
\label{tab1}
\end{center}
\end{table}

Following Table III, ConvLSTM + Met has the best RMSE which makes sense because meteorology has the most significant impact on air pollution. We also observe that the RMSE of ConvLSTM + All model is not the best. We can explain that a naive combination with the same weights for all factors will not be very efficient since in actual, air pollution could be impacted by other factors in complicated ways.

To evaluate how other spatiotemporal factors affect the air pollution interpolation’s efficiency, we propose to use the following test: removing one of the existing air pollution values from input data but still keep the values of other impacted spatiotemporal data and then check the regression error of the interpolated air pollution value with the existing one. If the error is small then we can infer that other spatiotemporal data has a remarkable effect on air pollution interpolation. To measure the error, we alternately set the air pollution value of each existed input pixel to zero, keep other data of that pixel unchanged, running the trained model on this modified input data and calculate RMSE between the inferred value with the actual pixel value of the same position. The final error is the mean of all errors after doing this procedure with all existing air pollution values. We call this error \textbf{spRMSE}, which means the RMSE caused by spatiotemporal factors. The experiments’ results are shown in Table III. We can notice that ConvLSTM + Speed model has a better \textbf{spRMSE} than ConvLSTM in spite it has worse overall RMSE; which indicates the driving average speed affects air pollution in spatiotemporal form. ConvLSTM + All model has the best \textbf{spRMSE}, proves that we can improve the citywide interpolation with more spatiotemporal data.

\subsection{Air pollution Forecasting: Experiments and Evaluations}
The baseline model for Air pollution forecasting is also a DAL forecasting model. This model has the same structure as the DAL model for interpolation but the input time steps are 24 hours and the prediction time lags are 12 hours. We still pre-train an Auto-Encoder and then use it to train the prediction model. The spatial loss is computed by summing up the spatial loss for each 12 output image frames. The temporal loss is also the sum of the loss between 1 image slice with 2 neighbor image slices of the output (the DAL paper chose temporal neighbor size is 2). The final loss is the total of labeled loss and spatiotemporal loss.

The forecasting ConvLSTM network also predicts 12 hours from 12 previous hours as the input time steps. The number of encoder layers is 3 as the same number for forecasting layers. The output channels are 16, 16 and 32, respectively.

Similar to the interpolation experiment section, we build 2 predicting models based on CNN Encoder-Decoder and FC-LSTM. The CNN Encoder-Decoder model has 3 layers for encoder and 3 layers for the decoder part which is similar to ConvLSTM predicting model.

Table IV shows the RMSE of experimental models in the test set. As expected, ConvLSTM model gives the best RMSE, following is the CNN Encoder-Decoder and the last 2 positions are DAL and FC-LSTM.

\begin{table}[htbp]
\caption{Prediction error (RMSE) of ConvLSTM and baseline models} 
\begin{center}
\begin{tabular}{|c|c|} 
    \hline \textbf{Models} & \textbf{RMSE} \\
    \hline \bfseries{ConvLSTM} & \bfseries{8.59883} \\
	\hline Deep Air Learning (DAL) \cite{b15} & 9.44042 \\ 
	\hline CNN Encoder-Decoder & 9.16437  \\ 
	\hline Stacked FC-LSTM & 21.22256  \\ 
	\hline
\end{tabular}
\label{tab1}
\end{center}
\end{table}

\subsubsection{Forecasting with air pollution influence factors}
For the next experiment, we evaluate the forecasting results in term of combination with air pollution related spatiotemporal factors. The examining models are: ConvLSTM (as baseline), ConvLSTM + Met (air pollution and meteorological data), ConvLSTM + Traffic (air pollution and transportation traffic data), ConvLSTM + Speed (air pollution and vehicles average speed data), ConvLSTM + External (air pollution and external air pollution data), ConvLSTM + All (air pollution and all related factors). Table V shows the RMSE of each examined models in the test set.

\begin{table}[htbp]
\caption{RMSE of ConvLSTM model with spatiotemporal factors}
\begin{center}
\begin{tabular}{|c|c|} 
    \hline \textbf{Models} & \textbf{RMSE} \\
	\hline ConvLSTM (baseline) & 8.59883 \\ 
	\hline \bfseries{ConvLSTM + Met} & \bfseries{8.43047} \\
	\hline ConvLSTM + Traffic & 8.53342 \\ 
	\hline ConvLSTM + Speed & 8.58124 \\ 
	\hline ConvLSTM + External & 8.53036 \\ 
	\hline ConvLSTM + All & 8.46117  \\ 
	\hline
\end{tabular}
\label{tab1}
\end{center}
\end{table}

Following Table V, the ConvLSTM + Met model has the best RMSE, and ConvLSTM + All model takes the second place. Therefore, we still strongly recognize the effect of spatiotemporal factors on air pollution forecasting problem.

\section{Related work}
Air pollution interpolation was started researching a few years ago. There are some papers such as in \cite{b11,b19} that try to interpolate Air pollution at locations where are lack of monitoring stations. They proposed to use some basic interpolation methods such as Spatial averaging, Nearest neighbor, Inverse distance weighting (IDW), Kriging \cite{b19}, and Shape Function based spatiotemporal interpolation \cite{b11}. By our evaluation, these are basic and simple interpolation models that often used as baselines for more advanced methods.

Recently, some air pollution related research has leveraged Machine Learning/Neural Networks based models in predicting Air pollution \cite{b3,b7,b15,b23,b24}. Most of these papers used common datasets like monitoring air pollution, meteorological data. Some papers used specific datasets such as GPS trajectories generated by over 30,000 taxis in Beijing in \cite{b23} or road networks data in \cite{b7}. The proposed air pollution predicting models are quite diversity, from co-training-based semi-supervised learning approach \cite{b23}, linear regression-based temporal predictor \cite{b24} to Spatiotemporal Semi-supervised Learning \cite{b15} and Attention Model \cite{b3}. Some research proposed grid-based air pollution interpolation or prediction. Nevertheless, they only focused on forecasting air pollution for discrete locations, not considering the whole city to be an image as in our approach. Furthermore, they used much hand-crafted spatial and temporal features that were difficult to generalize to other similar problems.

Furthermore, we survey general Spatiotemporal Deep Learning algorithms. In \cite{b16}, the authors have proposed a ConvLSTM model and used for precipitation forecasting. In the paper, the authors demonstrated that ConvLSTM was better than Fully Connected LSTM in spatiotemporal problems like moving MNIST and weather radar echo images of Hong Kong for precipitation forecasting.

The spatiotemporal problem is also fit for the crowd flows prediction problem. In \cite{b20}, the authors presented a Deep Neural Network Spatiotemporal (DeepST) for predicting crowd flows in Beijing and New York. They proposed the DeepST model based on Convolutional Neural Network and used Residual Units (as in ResNet model) to build a very deep network to capture more citywide dependencies. The last layer was a Fusion layer to combine deep network results with external factors (such as meteorology, holidays).

\section{Conclusion and Future work}
In this research, we have introduced 3 main contributions. First, we described and resolved the citywide scale Air Pollution Interpolation and Prediction problem by considering a whole city to be one image. Second, we pointed out and collected several spatiotemporal datasets, which have effects on air pollution throughout the city. Lastly, we proposed a spatiotemporal Deep Learning based model for citywide air pollution interpolation and prediction. We proved that the proposed model does not only work better than CNN and LSTM themselves in spatial and temporal features analysis but also outperforms state-of-the-art relevant models. The leverage of using other spatiotemporal factors gives us a powerful model in interpolating and forecasting air pollution over the city.

Our proposed method for air pollution problem is also suitable for other urban spatiotemporal based predictions such as traffic volume prediction or crowd flow forecasting. In the future, we will extend this spatiotemporal research on urban traffic volume and driving speed data to foresee traffic congestion and other urban relating problems.

\vspace{12pt}
\end{document}